\documentclass[journal,letterpaper]{IEEEtran}
\pdfoutput=1

\usepackage{amsmath,amsfonts,amssymb,bm,mathrsfs}

\usepackage[linesnumbered,ruled,vlined]{algorithm2e}

\SetNlSty{textbf}{\color{gray}}{}   % 行号样式
\SetAlgoNlRelativeSize{-1}
\SetNlSkip{0.6em}
\bstctlcite{BSTcontrol:NoDash}

\usepackage{array,booktabs,multirow,makecell,bigstrut}
\usepackage[table]{xcolor}
\definecolor{rowcolor}{rgb}{0.898,0.949,0.969}
\usepackage{graphicx}
\usepackage{tikz}
\usepackage{textcomp}
\usepackage{url}
\usepackage{verbatim}
\usepackage{pifont}
\usepackage{cleveref}
\Crefname{figure}{Fig.}{Figs.}
\def\etal{et al.}
\usepackage{soul}
\usepackage{xcolor}
\sethlcolor{yellow!50} % 浅黄色高亮
\usepackage{xcolor}

\soulregister\cite7
\soulregister\ref7
\soulregister\pageref7
\soulregister\Cref7
\soulregister\etal7
\soulregister\textcolor7
\soulregister\text7

\usepackage{enumitem}

\usepackage[switch,pagewise]{lineno}

\usepackage[section]{placeins}
\usepackage{dblfloatfix}
\usepackage{afterpage}

\begin{document}
% \pagewiselinenumbers
% \switchlinenumbers     % 两栏：行号自动在外侧（可留可去）
% \modulolinenumbers[1]  % 可选，保证每行都标（非必须）
% \linenumbers

\title{ST-Booster: An Iterative Spatiotemporal Perception Booster for Vision-and-Language Navigation in Continuous Environments}

%\author{Anonymous Authors}
\author{Lu Yue, Dongliang Zhou, Liang Xie, Erwei Yin$^*$, and Feitian Zhang$^*$ % <-this % stops a space
\thanks{L. Yue and F. Zhang are with the Robotics and Control Laboratory, the School of Advanced Manufacturing and Robotics, and the State Key Laboratory of Turbulence and Complex Systems, Peking University, Beijing, 100871, China. L. Yue is also with the Defense Innovation Institute, Academy of Military Sciences, Beijing, 100071, China; and Tianjin Artificial Intelligence Innovation Center, Tianjin, 300450, China.}
\thanks{D. Zhou is with Harbin Institute of Technology, Shenzhen, Xili University Town, Shenzhen, 518055, China.}
\thanks{L. Xie and E. Yin are with the Defense Innovation Institute, Academy of Military Sciences, Beijing, 100071, China, and Tianjin Artificial Intelligence Innovation Center, Tianjin, 300450, China.}
\thanks{Corresponding authors: E. Yin (email: yinerwei1985@gmail.com) and F. Zhang (email: feitian@pku.edu.cn).}   
%\thanks{A preliminary version appeared on arXiv \cite{yue2025st}.}
}

%\markboth{Journal of \LaTeX\ Class Files,Vol.14, No.8, August2021}%

%\IEEEpubid{0000--0000/00\$00.00\copyright2021 IEEE}

\maketitle

\begin{abstract}
Vision-and-Language Navigation in Continuous Environments (VLN-CE) requires agents to navigate previously unseen and continuous spaces based on natural language instructions. Compared to discrete settings, VLN-CE poses two core perception challenges. First, the absence of predefined observation points leads to heterogeneous visual memories and weakened global spatial correlations. Second, cumulative reconstruction errors in three-dimensional scenes introduce structural noise, impairing local feature perception.
To address these challenges, this paper proposes ST-Booster, an iterative spatiotemporal booster that enhances navigation performance through multi-granularity perception and instruction-aware reasoning. ST-Booster consists of three key modules --- Hierarchical SpatioTemporal Encoding (HSTE), Multi-Granularity Aligned Fusion (MGAF), and Value-Guided Waypoint Generation (VGWG). HSTE encodes long-term global memory using topological graphs and captures short-term local details via grid maps. MGAF aligns these dual-map representations with instructions through geometry-aware knowledge fusion. The resulting representations are iteratively refined through  pretraining tasks. 
During reasoning, VGWG generates Guided Attention Heatmaps (GAHs) to explicitly model environment-instruction relevance and optimize waypoint selection. Extensive comparative experiments and performance analyses are conducted, demonstrating that ST-Booster outperforms existing state-of-the-art methods, particularly in complex, disturbance-prone environments. Our code is available at https://github.com/yueluhhxx/ST-Booster.
\end{abstract}

% \begin{NTP}
% Vision-and-Language Navigation in Continuous Environments demands robust perception to navigate reconstructed real-world scenes under observational noise. 
% This paper presents ST-Booster, a practical solution that enhances scene understanding by combining global layout tracking with local detail mapping. ST-Booster maintains fused spatiotemporal environmental representations by integrating navigation memory with real-time fine-grained observations, which are further aligned with linguisitc instructions. It autonomously identifies instruction-relevant regions via predicted heatmaps to inform path planning decisions. This multi-scale perception strategy significantly improves navigation robustness in noisy, cluttered indoor environments. 
% To support real-world applications, we open-source the code of ST-Booster, enabling its deployment in indoor autonomous navigation tasks for robotics practitioners.
% \end{NTP}


\section{Introduction}
\begin{figure}[!t]
  \centering
\includegraphics[width=1.0\linewidth]{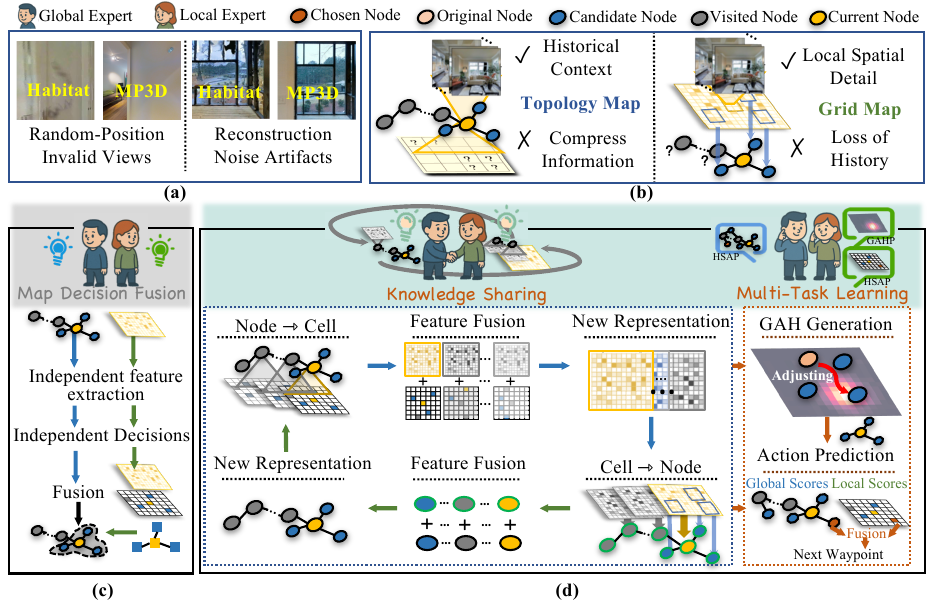}
    \vspace{-3mm}
   \caption{{Overview of environmental representation and perception in VLN-CE. (a) Comparison of observations between discrete environments (MP3D) and continuous environments (Habitat); (b) advantages and limitations of topological~\cite{chen2021topological} and grid maps~\cite{georgakis2022cross}; (c) workflow of BEVBert~\cite{an2023bevbert} using dual-map decision fusion; (d) workflow of ST-Booster with iterative dual-map feature fusion and multi-task decision integration.}\label{fig:motivation}}
   \vspace{-5mm}
\end{figure}

\IEEEPARstart{V}{ision}-and-Language Navigation (VLN)~\cite{anderson2018vision, tan2025source, wu2021improved} is an embodied intelligence task {where agents follow natural-language instructions to navigate using visual inputs from limited data~\cite{wang2024vision, wang2023res}.}
Traditionally, VLN frameworks operate within discrete environments, where robots move only between predefined waypoints. In contrast, VLN in continuous environments (VLN-CE)~\cite{krantz2020beyond,anderson2021sim} requires low-level control in continuous space.
Although VLN-CE offers a more realistic scenario for navigation tasks, it also introduces substantial challenges. 
{As illustrated in \Cref{fig:motivation}(a), without a predefined graph structure, the agent reasons over long horizons and handles noisy observations from challenging viewpoints at arbitrary poses in three-dimensional (3D) reconstructed environments. This requires stronger spatiotemporal perception and instruction alignment.}

Current approaches for VLN-CE are broadly categorized into end-to-end and map-based methods. End-to-end methods implicitly learn environmental perceptions through {RGB-D} image processing~\cite{hong2021vln, chen2021history, hong2024navigating}, facing challenges in preserving fine-grained spatial details and long-term history due to feature compression. Map-based methods as illustrated in \Cref{fig:motivation}(b) have emerged as solutions for achieving effective environmental perception, owing to their inherent advantages in structured representation. Particularly, topological mapping techniques~\cite{chen2021topological, huo2023geovln} address long-term dependencies by constructing graph representations; however, they often oversimplify the local representations. Conversely, grid-based methods~\cite{wang2023gridmm} excel in capturing detailed local features but suffer from high computational cost for large-scale environments.
Recent hybrid strategies utilize both complementary map representations. Initial efforts~\cite{liu2023bird} directly combined two 
map features through averaging operations, which diluted critical spatial relationships. Subsequent advancements~\cite{an2023bevbert} introduced multi-scale fusion mechanisms (\Cref{fig:motivation}(c)). However, the prevailing late-fusion paradigm inherits limitations from individual map representations.

\begin{figure*}[t]
  \centering
    \includegraphics[width=0.98\linewidth]{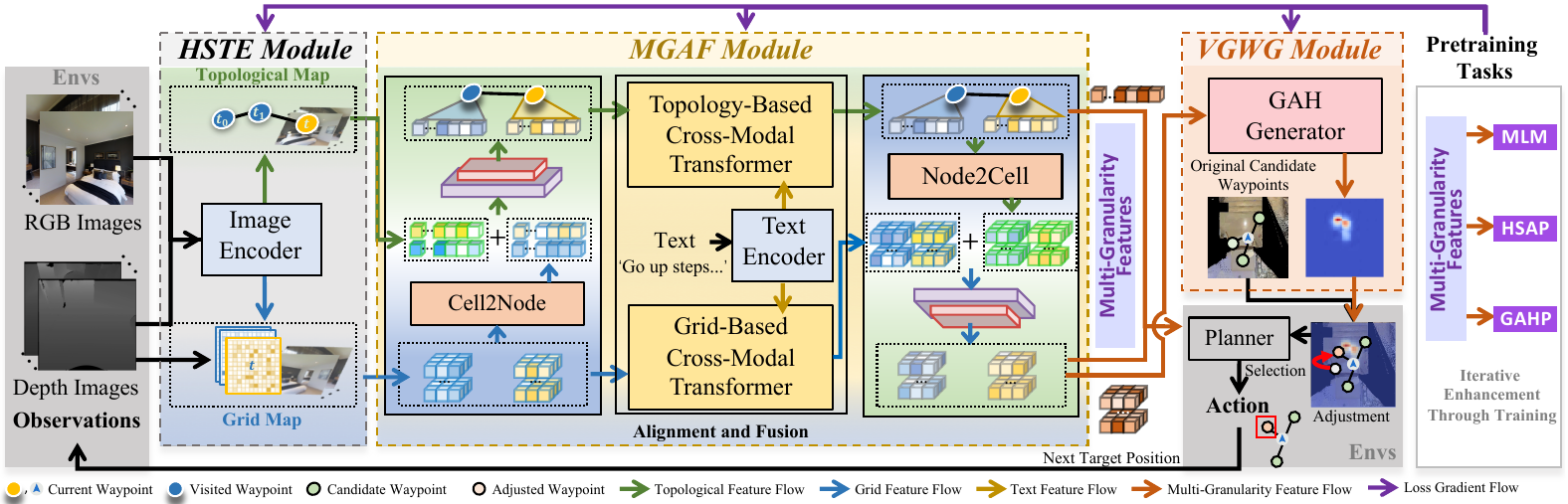}
    %\vspace{-10mm}
   \caption{The proposed ST-Booster comprises three core modules. First, the Hierarchical SpatioTemporal Encoding (HSTE) module captures global spatial structures via global topological graphs and extracts local temporal details using local grid-based maps. Next, the Multi-Granularity Aligned Fusion (MGAF) module fuses these heterogeneous map features and aligns the fused representations with linguistic embeddings. Finally, the integrated representations are utilized in the Value Guided Waypoint Generation (VGWG) to predict Guided Attention Heatmaps (GAHs), which adaptively adjust candidate waypoint distributions to instruction-relevant regions.}
   \label{fig:method}%
\end{figure*}

{
To tackle the perception challenges in VLN-CE, we introduce ST-Booster, an iterative spatiotemporal enhancement framework built upon the dual-map knowledge sharing idea (\Cref{fig:motivation}(d)). ST-Booster consists of three main modules (\Cref{fig:method}).
First, the Hierarchical SpatioTemporal Encoding (HSTE) module balances long- and short-term spatiotemporal dependencies by constructing global topological maps for long-term reasoning and local grid maps for detailed perception. Although these maps are complementary, their heterogeneous representations hinder effective feature-level transfer. To address this, we propose the Multi-Granularity Aligned Fusion (MGAF) module, which fuses maps bidirectionally by adding global context to grids and local details to topological nodes for unified semantic understanding. Finally, to enhance interpretability and strengthen decision-level reasoning, we include the Value-Guided Waypoint Generation (VGWG) module, which predicts Guided Attention Heatmaps (GAHs) to highlight instruction-relevant regions and generate meaningful candidate waypoints.
Extensive experiments demonstrate that ST-Booster achieves robust and consistent navigation in noisy and complex environments, effectively integrating spatial and temporal cues for more reliable perception and decision making.}

{
The contributions of this paper are threefold. First, 
this paper introduces ST-Booster, an iterative refinement framework that integrates the advantages of global topological and local grid maps in perceiving spatiotemporal information at the feature space level. 
Second, this paper designs a weakly supervised module that predicts Guided Attention Heatmaps (GAHs) to focus candidate waypoints on instruction-relevant regions, improving decision robustness and interpretability.
Third, extensive experiments on VLN-CE validate the effectiveness of the proposed framework, demonstrating consistent performance gains over state-of-the-art methods, particularly in unseen and noisy environments.}

% The remainder of this paper is organized as follows. Section \ref{related work} conducts a comprehensive review of the relevant literature. The design of ST-Booster is presented in Section \ref{stbooster}. Section \ref{experiment} details dataset construction and presents the experimental results. Finally, Section \ref{conclusion} provides the concluding remarks.

\section{Related Work}\label{related work}
In this section, we review related research on VLN-CE and environment representation in autonomous navigation. Comparison with existing literature is carried out to highlight the novel aspects of our research.

\noindent\textbf{Vision-and-Language Navigation in Continuous Environments (VLN-CE).}
Facing the challenges of VLN-CE, researchers have explored various techniques.
Initial methods attempted to predict low-level actions from continuous visual observations through end-to-end training~\cite{krantz2020beyond, raychaudhuri2021language, zhao2023mind}. 
However, these methods struggled to capture long-term dependencies along paths with overlapping observations.
To address this limitation, researchers introduced waypoint predictors, which leverage predicted local navigation graphs to provide candidate waypoints for subsequent decision-making~\cite{krantz2021waypoint, hong2022bridging, krantz2022sim}.
This modular approach bridges the gap between discrete and continuous environments and has been extended with semantic information~\cite{wu2024vision, an2021neighbor, georgakis2022cross, chen2022weakly,irshad2021sasra}.
Nevertheless, VLN-CE performance remains below discrete VLN~\cite{li2023improving, wang2023dreamwalker}.
A key reason is that both end-to-end and modular approaches rely on implicit visual representations, hindering spatiotemporal reasoning and vision-textual dependencies.
{
To enhance structured spatiotemporal reasoning, we extend the dual-map expert framework of An~\etal~\cite{an2023bevbert} with an iterative spatiotemporal perception booster that performs feature-level fusion and cross-map knowledge transfer. This mechanism integrates complementary global and local cues into a complete scene representation. Furthermore, a multi-task joint training scheme reinforces the fused features and provides instruction-aware priors for sampling multimodally aligned waypoint candidates, improving both navigation accuracy and interpretability.
}
% Moreover, the hybrid-map representation provides coarse-grained prior knowledge, offering candidate waypoints with multi-modal alignment cues.

\noindent\textbf{Environment Representation.}
Scene perception is essential for navigation, offering agents visual cues and contextual insights~\cite{min2021film, duh2020v}. 
Traditional end-to-end training methods often lack structured constraints, leading to loss and distortion of representation.
To address this issue, researchers have explored map-based approaches to effectively manage perceptual data~\cite{huang2023visual, jiang2024bevnav}. 
For instance, Chen~\etal ~\cite{chen2021topological} introduced a framework combining topological maps and attention for planning. In a later study~\cite{chen2024mapgpt}, an online topological map further improves spatial understanding.
Some other approaches leverage grid maps to represent environmental details. Unlike topological graphs that compress visual representations into nodes, grid maps offer detailed local spatial information, aiding short-term decision-making. For instance, 
Georgios~\etal ~\cite{georgakis2022cross} proposed a cross-modal map learning framework to enhance path prediction. Meanwhile, Huang~\etal ~\cite{huang2023visual} integrated pre-trained vision-language features with 3D reconstruction for spatial mapping. 
To combine their strengths, hybrid map methods have been proposed. For instance, Liu~\etal ~\cite{liu2023bird} constructed a global topological map by averaging grid map features and fused the decision results. 
In contrast, An~\etal ~\cite{an2023bevbert} enhanced both maps by incorporating panoramic image features and their low-dimensional representations, improving action reasoning via decision fusion. 
However, existing hybrid map methods are still constrained by the inherent limitations of each map type without feature-level fusion. 
This paper proposes an iterative spatiotemporal perception booster that focuses on feature-level hybrid map fusion to achieve more comprehensive, multi-granularity perception.

\section{ST-Booster}\label{stbooster}
In this section, we present the problem formulation of VLN-CE (\Cref{sec:problem}) and introduce ST-Booster, an iterative spatiotemporal framework for perception in complex environments. As illustrated in \Cref{fig:method}, the HSTE module builds hybrid maps by combining global topological and local grid representations (\Cref{sec:perception}), which are fused and aligned with instructions through the MGAF module (\Cref{sec:framework}). The resulting multi-granularity features are used by the VGWG module to predict GAHs for instruction-relevant waypoint generation (\Cref{sec:GAH}) and are iteratively refined through pretraining and fine-tuning with online-generated maps for navigation reasoning (\Cref{sec:fine}).

\subsection{Problem Formulation}
\label{sec:problem}
Following the conventional setting of VLN-CE~\cite{an2024etpnav, an2023bevbert}, an agent receives an instruction containing $L$ words for each navigation episode.
This instruction is embedded as $\mathbf{I} = \{{\bm{l}_1, \cdots, \bm{l}_L}\}$, where $\bm{l}_i$ represents the embedding of the $i$-th word.
During the navigation process, the agent records its position information and observes the surrounding panoramic RGB-D images $\mathbf{O}_t=\{(\mathbf{o}^{\text{rgb}}_{t,i}, \mathbf{o}^{\text{d}}_{t,i})\}_{i=0}^{11}$ at each time step $t$.
These images are taken from 12 distinct views, each oriented at a horizontal heading angle of $\theta_i = 30^{\circ}\times i$, where $i=0,\cdots,11$. Here, $\mathbf{o}^{\text{rgb}}_{t,i}\in \mathbb{R}^{H\times W\times 3}$ and $\mathbf{o}^{\text{d}}_{t,i}\in \mathbb{R}^{H'\times W'}$ represent
the RGB and depth images observed from the $i$-th view, respectively. 
The agent's objective is to effectively integrate textual and visual information to navigate toward the target locations specified in the instruction.
In contrast to selecting discrete waypoints, agents in VLN-CE navigate within a 3D continuous environment by choosing actions from a low-level action space defined as 
$\mathscr{A}=\{\text{forward 0.25m}, \text{turn left } 15^{\circ}, \text{turn right } 15^{\circ}, \text{stop}\}$.

%\subsection{Dual-Map spatiotemporal Perception Module}\label{sec:perception}
\subsection{Hierarchical SpatioTemporal Encoding Module}\label{sec:perception}
To effectively navigate complex continuous environments, we propose a hierarchical spatiotemporal encoding module that decomposes the environmental representation into long-term and short-term components.

%\noindent\textbf{Topological Maps for Long-Term Perception.}
\noindent\textbf{Structural Topology Encoding for Global Perception.}
At step $t$, the global topological map $\mathcal{G}_t = \{\mathcal{V}_t, \mathcal{E}_t\}$ captures long-term environmental awareness by jointly modeling visited and observed waypoints. The node set $\mathcal{V}_t = \mathcal{V}^\text{v}_t \cup \mathcal{V}^\text{o}_t$ stores the visited nodes $\mathbf{n}\in \mathcal{V}^\text{v}_t$ with averaged panoramic visual representations and the observed nodes $\mathring{\mathbf{n}}\in \mathcal{V}^\text{o}_t$ with view-specific visual representations, where $\mathcal{V}^\text{v}_t = \{\mathbf{n}_i\}_{i=1}^{t}$ and $\mathcal{V}^\text{o}_t = \{\{\mathring{\mathbf{n}}_{i, k}\}_{k=1}^{g_i}\}_{i=1}^t$, $g_i$ is the number of observed waypoints at $i$-th time step. In particular, the visual representations include the RGB features extracted via ViT~\cite{dosovitskiy2020image} and the depth features extracted from ResNet~\cite{he2016deep}.
In addition, the edge set $\mathcal{E}_t$ encodes the Euclidean distances between adjacent nodes. The nodes in $\mathbf{G}_t$ dynamically integrate perception context through fused relative position embeddings and time-step embeddings. Additionally, a globally connected virtual node $\mathbf{n}_0$ is created to represent the `stop' action.
Topological maps efficiently manage global scaling and maintain historical memory, but compress the representation details.
To address this, we integrate local grid maps to enrich fine-grained environmental representation.

%we represent long-term perception using a topological map $\mathcal{G}_t = \{\mathcal{V}_t, \mathcal{E}_t\}$.
%This graph structure tracks both visited and observed waypoints along the navigation path.
%In particular, node set $\mathcal{V}_t$ stores the node observations and is partitioned into visited and unvisited subsets: $\mathcal{V}_t = \mathcal{V}^\text{v}_t \cup \mathcal{V}^\text{o}_t$}, while edge set $\mathcal{E}_t$ encodes Euclidean distances between adjacent nodes. 
%We extract visual features utilizing a pre-trained vision transformer (ViT)~\cite{dosovitskiy2020image} for RGB images and a pre-trained residual network~\cite{he2016deep} for depth image processing in point-goal navigation~\cite{wijmans2019dd}.
%The framework maps the averaged panoramic features to visited and current nodes in $\mathcal{V}^\text{v}_t$.
%For nodes observed but not yet visited in $\mathcal{V}^\text{o}_t$, visual representations are recorded as specific view features.
%The node representations are further updated to capture spatial and temporal dependencies through the integration of relative position embeddings and time step embeddings.
%For simplicity, we denote all node representations at step $t$ as $\mathbf{n}_t$, which together form the topological map $\mathbf{G}_t$.
%As for the `stop' action, a virtual node $\mathbf{n}_0$ is created, which is connected to all nodes. 
%Topological maps efficiently manage global scaling and maintain historical memory but compress the representation details.
%To address this, we integrate grid maps to enrich fine-grained environmental representation.

\afterpage{%
  \FloatBarrier
\begin{figure}[!t]
  \centering
   \includegraphics[width=1\linewidth]{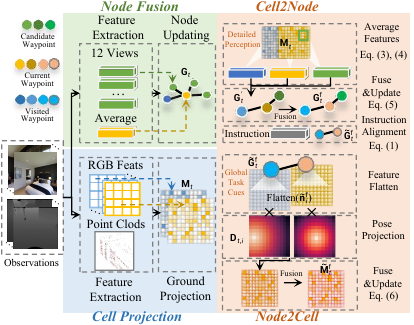}
   % \caption{The feature fusion approach in this paper. Visual observations are stored in both topological nodes and grid cells. For interaction between two feature spaces, we flatten node features according to the global spatial relationships, thereby enriching the grid features with historical context. Additionally, grid feature aggregation is used to supply fine-grained spatial information to the topological nodes.}
   \vspace{-5mm}
   \caption{{Illustration of the proposed feature fusion approach. Visual observations are embedded in topological nodes and grid cells. Node features are flattened by global spatial relations to add historical context to grid features, while grid aggregation supplies fine-grained spatial details to nodes.}}
   \label{fig:feat}
   
\end{figure}
}

%\noindent\textbf{Grid Maps for Short-Term Perception.}
\noindent\textbf{Dense Grid Encoding for Local Perception.}
{To represent the detailed surroundings of the agent, we construct a local grid map $\mathbf{M}_t \in \mathbb{R}^{U \times V \times D}$ centered on the agent, where $U$ and $V$ are spatial dimensions and $D$ is the feature dimension. 
To reduce computational cost while maintaining geometric fidelity, $\mathbf{M}_t$ is built from dimensionality-reduced RGB features and aligned depth inputs. Specifically, patch-level RGB embeddings $\mathbf{V}^{\text{M-R}}_t \in \mathbb{R}^{12 \times P \times P \times D}$ are extracted using a ViT encoder~\cite{dosovitskiy2020image} with patch size $P$. The corresponding depths ${\mathbf{o}^{\text{d}}_{t,i}}, i=0,1,\ldots,11$ are average-pooled to the same $P \times P$ layout, denoted as $\mathbf{d}_t = \text{AvgPool}({\mathbf{o}^{\text{d}}_{t,i}})$. Each depth value in $\mathbf{d}_t$ is then back-projected into 3D and transformed into the ego-centric frame using the current pose, forming a point cloud $\mathbf{P}_t = [\mathbf{p}_{t,1}, \dots, \mathbf{p}_{t,N}]$ with $N = 12\times P \times P$ points. Each 3D point $\mathbf{p}_{t,i}$ inherits its RGB embedding $\mathbf{V}^{\text{M-R}}_{t,i}$, producing semantically enriched point features.
To obtain a compact spatial representation, the point features are projected onto a Bird’s-Eye-View (BEV) grid of size $U \times V$. Invalid or out-of-range points are removed, and valid points are assigned to grid cells based on their coordinates. For each grid cell $\mathbf{m}_{(u,v)}$, the aggregated feature is computed as $\mathbf{m}_{(u,v)} = \frac{1}{|\mathcal{S}_{(u,v)}|}\sum_{i\in \mathcal{S}_{(u,v)}}\mathbf{V}^{\text{M-R}}_{t,i}$, where $\mathcal{S}_{(u,v)}$ denotes the set of points falling into that cell. This process converts the sparse 3D point cloud into a structured BEV feature map $\mathbf{M}_t$, providing a compact spatial representation for downstream reasoning. For each grid cell $\mathbf{m}_{(u,v)}$, positional embeddings are added to encode spatial relationships between neighboring cells. The resulting grid map offers fine-grained local environmental information, complementing global representations.}

\subsection{Multi-Granularity Aligned Fusion Module}\label{sec:framework}
%This section presents a collaborative enhancement of the two maps obtained in \Cref{sec:perception}, aligning them with instruction cues to refine the multi-modal hybrid map representation. 
%We introduce two domain-specific multi-modal matching operators and a cross-domain knowledge-exchange operator to enable instruction intent alignment and cross-domain knowledge sharing, respectively. During iterative training, these operators are alternately activated to strengthen the multi-modal hybrid map perception.
Although global topological and local grid maps offer complementary perceptual strengths, their structural limitations still hinder unified scene understanding. To overcome this, we adopt an asymmetric fusion strategy within a multi-granularity aligned fusion module that performs instruction feature alignment and cross-domain perception fusion to bridge heterogeneous long- and short-term perceptions. In this design, topological features are first enriched with fine-grained grid information before instruction alignment, whereas grid features are aligned with instructions first and then fused with globally aligned task cues to preserve local geometric detail while incorporating global semantics.

\noindent\textbf{Instruction Feature Alignment.}\label{sec:text-align}
By interacting with instruction information, contextual understanding is embedded into domain-specific environmental perception. We adopt Topology-based Cross-Modal Transformer (TCMT) and Grid-based Cross-Modal Transformer (GCMT) ~\cite{an2023bevbert} to facilitate the fusion of instruction cues with long-term and short-term environmental perception, respectively. 
Specifically, TCMT incorporates a graph-aware self-attention mechanism to integrate instructions $\mathbf{I}$ with the global topological representation enhanced by grid features ($\mathbf{G}^\text{f}_{t}$ introduced in \Cref{eq:GF}), yielding a globally informed multi-modal graph $\tilde{\mathbf{G}}^{\text{f}}_{t}$, i.e.,
\begin{linenomath*}
\begin{equation}\label{eq:TCMT}
    \tilde{\mathbf{G}}^{\text{f}}_{t}=\text{TCMT}(\text{NE}(\mathbf{{G}}^{\text{f}}_{t}), \mathbf{I}),
\end{equation}
\end{linenomath*}
where NE$(\cdot)$ represents the node embedding network. 
For GCMT, the fusion of instruction information $\mathbf{I}$ and local spatial awareness $\mathbf{M}_{t}$ is achieved through a self-attention mechanism for cell encoding, i.e.,
\begin{linenomath*}
\begin{equation}\label{eq:GCMT}
    \tilde{\mathbf{M}}_{t}=\text{GCMT}(\text{CE}(\mathbf{M}_{t}), \mathbf{I}),
\end{equation}
\end{linenomath*}
where $\tilde{\mathbf{M}}_{t}$ represents the local-aware multi-modal grid maps and CE$(\cdot)$ represents the grid cell embedding network.

%We integrate instructional information within each domain to enable the topology-based maps to comprehend the context from the global environment and the grid-based maps to capture instructional cues from local details. However, the knowledge across domains remains independent, hindering the agent’s ability to form a comprehensive spatiotemporal understanding of the environment. To address this, we introduce cross-domain knowledge transfer to enhance the integration of hybrid map perception.
This dual-stream design establishes hierarchical perception, where global topological graphs comprehend navigation context through global trajectory-linguistic correlation, while local grid maps discern fine-grained spatial-textual correspondence. However, this domain alignment creates independent representations that impede unified spatiotemporal reasoning. 
%To address this, we introduce geometric-constrained cross-domain interaction to bridge these complementary perspectives.

\noindent\textbf{Cross-Domain Perception Fusion.}
To leverage the complementary strengths of both map structures, we introduce a geometrically-constrained projection operator to facilitate cross-domain knowledge sharing.
The operator comprises two components,  \textit{Cell2Node} and \textit{Node2Cell}, which project perceptual information from heterogeneous domains into a unified feature space.
As depicted in the upper-right corner of \Cref{fig:feat}, \textit{Cell2Node} is initially utilized to transfer local grid features into the long-term topological domain. To maintain a domain-invariant structure, suitable feature migration strategies are applied based on the different node types.
In particular, grid maps are aggregated to provide comprehensive environmental information for the visited nodes $\mathbf{n}$. As for the unvisited nodes $\mathring{\mathbf{n}}$ with partly observations, we locate them on the grid cells and extract the corresponding features. 
With the utilization of \textit{Cell2Node}, the intermediate topological map $\mathbf{G}'_t$ is generated, consisting of the visited nodes $\mathbf{n}'$ and observed nodes $\mathring{\mathbf{n}}'$ with new representations from $\mathbf{M}_t$ as follows,
% \begin{equation}
%     \mathbf{n}'_t = \left\{
%     \begin{aligned}
%         & \text{average}(\mathbf{M}_t),
%         \textrm{if}\,\mathbf{n}_t \ \text{is the visited node}, \\
%         & \mathbf{M}_t[x_t, y_t],
%         \textrm{if}\,\mathbf{n}_t  \ \text{is the observed node},
%     \end{aligned}s
%     \right.
% \end{equation}
\begin{linenomath*}
\begin{equation}
    \mathbf{n}_t' = \text{average}(\mathbf{M}_t), 
\end{equation}
\begin{equation}
    \mathring{\mathbf{n}}'_{t, i} = \mathbf{m}_{(u_{t, i}, v_{t, i})}, i=1, \cdots, g_t,
\end{equation}
\end{linenomath*}
where average$(\cdot)$ represents an expectation pooling operation, $u_{t, i}$ and $v_{t, i}$ denote the geometric coordinates of the $i$-th observed node at time step $t$ when projected onto the grid map, $g_t$ is the observed candidate number and $\mathbf{m}_{(u_{t, i}, v_{t, i})}$ represents the grid cell features with the coordinate $(u_{t, i}, v_{t, i})$ in grid map $\mathbf{M}_t$.
Subsequently, $\mathbf{G}'_t$ is fused with the original graph $\mathbf{G}_t$ to form a hybrid graph with complete perception, i.e.,
\begin{linenomath*}
\begin{equation}\label{eq:GF}
    \mathbf{G}^{\text{f}}_{t}=\text{GF}([\mathbf{G}'_{t}; \mathbf{G}_t]), 
\end{equation}
\end{linenomath*}
where $[\cdot;\cdot]$ denotes the concatenation operation, and GF$(\cdot)$ is a fully-connected network fusing the  integrated graph features.
Subsequently, the fine-grained enhanced $\mathbf{G}^{\text{f}}_{t}$ is aligned with the instruction in \Cref{eq:TCMT} to obtain $\tilde{\mathbf{G}}^{\text{f}}_{t}$.

Similarly, we introduce $\textit{Node2Cell}$ to project the historical perception of nodes into the local grid-based domain, which is illustrated in the bottom-right corner of \Cref{fig:feat}. 
To effectively capture comprehensive global context, such as task progress, we leverage  the enhanced linguistically-aligned topological maps $\tilde{\mathbf{G}}^{\text{f}}_t$. Their node features $\tilde{\mathbf{n}}^{\text{f}}_t$ are sequentially processed by \Cref{eq:GF} and \Cref{eq:TCMT} to facilitate cross-domain knowledge fusion.
Specifically, we flatten the node features and map them back into the grid maps with the same geometric positions. To reduce perceptual bias from isolated node features, we fuse multiple nodes within the grid map’s spatial boundaries, represented as $\mathcal{N}_t = \{\tilde{\mathbf{n}}^{\text{f}}_{t,i}|\text{dis}(\tilde{\mathbf{n}}^{\text{f}}_t, \tilde{\mathbf{n}}^{\text{f}}_{t,i})\text{\textless} U_b\}, \tilde{\mathbf{n}}^{\text{f}}_{t,i} \in\mathcal{V}^{\text{v}}_t$, where dis$(\cdot,\cdot)$ calculates the Euclidean distance between two nodes, and $U_b$ represents the geometry size of the grid maps. We also introduce a geometric-aware discount matrix $\mathbf{D}_{t,i}$ to adjust the contribution of multiple projected node features to each grid cell. Specifically, the nodes in $\mathcal{N}_t$ are projected onto the current grid map with coordinates $(u_{t,i}, v_{t,i}), i=1,\dots, \text{len}(\mathcal{N}_t)$. We calculate the Euclidean distances $d_{t,i,n}$ $(n = 0,\dots, U\times V - 1)$ between each grid cell and the grid at the map center, where $d_{t,i}^{\text{min}}$ and $d_{t,i}^{\text{max}}$ denote the minimum and maximum distances, respectively. Through the distance normalization, the discount matrix $\mathbf{D}_{t,i}$ is obtained by $\mathbf{D}_{t,i} = \{\frac{d_{t,i}^{\text{max}} - d_{t,i, n}}{d_{t,i}^{\text{max}} - d_{t,i}^{\text{min}}}\}$.
The intermediate perception is further updated from the multi-modal grid maps $\tilde{\mathbf{M}}_t$ obtained in \Cref{eq:GCMT}, i.e.,
\begin{linenomath*}
\begin{equation}
    \tilde{\mathbf{M}}^{\text{f}}_{t}=\text{MF}([(\sum_i\mathbf{D}_{t,i}\times \text{flatten}(\tilde{\mathbf{n}}^{\text{f}}_{t,i})); \tilde{\mathbf{M}}_t]),
\end{equation}
\end{linenomath*}
where MF$(\cdot)$ serves as a fusion network of grid map features.
Through bidirectional geometric-aware transformation, topological graphs are enriched with geometric details, grid maps are enhanced with global context, and multi-granularity map representations are constructed.

\subsection{Value-Guided Waypoint Generation Module}
\label{sec:GAH}
We introduce the Guided Attention Heatmap (GAH) prediction task to enhance model interpretability by explicitly aligning hybrid map perception with navigation instructions.
GAH is a self-centered heatmap that predicts the 2D probability distribution of instruction-relevant regions from multi-granularity grid map representations $\tilde{\mathbf{M}}^{\text{f}}_{t}$,
\begin{linenomath*}
\begin{equation}\label{eq:gah}
\mathbf{H}_{t} = \textrm{HFFN}(\tilde{\mathbf{M}}^{\text{f}}_{t}),
\end{equation}
\end{linenomath*}
where HFFN$(\cdot)$ denotes a fully connected forward network.
The ground-truth heatmap is derived from the next target waypoint in the reference trajectory, projected onto an egocentric coordinate system aligned with the agent’s heading and smoothed using a Gaussian kernel. To better capture detailed spatial relationships, the map resolution is upsampled by $m\times n$, resulting in a $(mU)\times (nV)$ high-definition grid. Based on these ground-truth GAHs, a weakly supervised training scheme (see \Cref{sec:fine}) is used to help the model learn contextual cues from hybrid multimodal perception.

{We integrate GAH as a plug-and-play module in the evaluation phase. As illustrated in \Cref{algorithm:GAH}, ST-Booster consists of two stages: prior guidance and fusion-based decision making. In the first stage, the waypoint predictor WP$(\cdot)$ generates an initial distribution $\mathbf{P}_t$, which is refined through weighted fusion with the predicted $\mathbf{H}_t$ from Eq.~(\ref{eq:gah}). Superscripts in \Cref{algorithm:GAH} indicate features of different stages (e.g., `s1' and `s2' denote the first and second stages, respectively) but are omitted here for clarity. $\mathbf{H}_t$ adaptively adjusts the heatmap weights, guiding candidate waypoints $\hat{\mathbf{w}}_{t,k}$ toward instruction-relevant regions. In the second stage, the model performs joint global–local reasoning based on the refined candidate distribution and fused features. It maintains long-term consistency through global topological cues while optimizing spatial precision using local grids, and finally selects the target waypoint from both current and historical candidates.}

\subsection{Pretraining and Fine-Tuning}\label{sec:fine}

\noindent\textbf{Pretraining.}
To enhance the representational capability of the hybrid multi-modal maps, we introduce three pretraining tasks.

\begin{algorithm}[!t]
    \caption{VLN-CE agent evaluation with GAHs.}
    \small
    \label{algorithm:GAH}
    \small

    % % 禁用行号
    % \SetAlgoNoLine
    \KwIn{Pre-trained `HFFN$(\cdot)$' in \Cref{sec:GAH}; fine-tuned agent `ST-Booster' in \Cref{sec:framework}; waypoint predictor `WP$(\cdot)$'; vision encoder `Encoder$(\cdot)$'~\cite{dosovitskiy2020image}, ~\cite{he2016deep}; map update operation `Update$(\cdot)$' in \Cref{sec:perception}; hybrid action predictor `AP$(\cdot)$' in \Cref{eq:action_score} and \Cref{eq:action_fuse} of \Cref{sec:fine}; add new observation to history path operation `Add$(\cdot)$'; environmental operation `Execute$(\cdot)$'; instruction embeddings $\mathbf{I}$; initial Observation $\mathbf{O}_0$; initial path $\mathbf{P}_0={\mathbf{O}_0}$; initial maps $\mathbf{G}_{0}$ and  $\mathbf{M}_{0}$; max time step $t_m$ and GAH weight $\delta$.}
    \KwOut{Generated path $\mathbf{P}_{T}$.}
    
    % \SetAlgoLined
    \For{$t = 1$ \KwTo $T\leq t_m$ }{
    
        $\mathbf{V_t}$ = Encoder($\mathbf{O}_{t}$);

        $\mathbf{P}_t = \text{WP}(\mathbf{V_t})$;
        
        $\mathbf{w}_{t,k} = \text{Sample}(\mathbf{P}_t), k=1,..., g_t$;\textcolor[rgb]{0.5,0.5,0.5}{ \% Sample original candidate waypoints}
        
        {$\mathbf{G}^{\text{s1}}_t, \mathbf{M}^{\text{s1}}_t = \text{Update}(\mathbf{G}_{t-1}, \mathbf{M}_{t-1}, \mathbf{w}_{t,k})$;}%\textcolor[rgb]{0.5,0.5,0.5}{ \% Update maps}

        {$\tilde{\mathbf{G}}^{\text{f-s1}}_t, \tilde{\mathbf{M}}^{\text{f-s1}}_t = \text{ST-Booster}(\mathbf{G}^{\text{s1}}_t, \mathbf{M}^{\text{s1}}_t, \mathbf{I})$;}

        \noindent{$\mathbf{H}_t=\text{HFFN}(\tilde{\mathbf{M}}^{\text{f-s1}}_t)$;\textcolor[rgb]{0.5,0.5,0.5}{ \% Generate GAHs}}

        $\hat{\mathbf{H}}_t=\delta\cdot\mathbf{H}_t + \mathbf{P}_t$; \textcolor[rgb]{0.5,0.5,0.5}{ \% Weighted fusion}

        $\hat{\mathbf{w}}_{t,k} = \text{Sample}(\hat{\mathbf{H}}_t), k=1,..., g'_t$; \textcolor[rgb]{0.5,0.5,0.5}{ \% Sample adjusted candidate waypoints}

        {$\mathbf{G}^{\text{s2}}_t, \mathbf{M}^{\text{s2}}_t = \text{Update}(\mathbf{G}_{t-1}, \mathbf{M}_{t-1}, \hat{\mathbf{w}}_{t,k})$;}%\textcolor[rgb]{0.5,0.5,0.5}{\% Re-update maps with $\hat{\mathbf{w}}^t_k$}

        $\tilde{\mathbf{G}}^{\text{f-s2}}_t, \tilde{\mathbf{M}}^{\text{f-s2}}_t = \text{ST-Booster}(\mathbf{G}^{\text{s2}}_t, \mathbf{M}^{\text{s2}}_t, \mathbf{I})$;

        $\mathbf{a}_t = \text{AP}(\tilde{\mathbf{G}}^{\text{f-s2}}_t, \tilde{\mathbf{M}}^{\text{f-s2}}_t)$;

        \If{$\mathbf{a}_t \rightarrow  \text{stop} $} {Break;}
        
        $\mathbf{O}_{t+1}$ = Execute($\mathbf{a}_{t}$); \textcolor[rgb]{0.5,0.5,0.5}{ \% Execute action in the environment}
        
        $\mathbf{P}_{t+1} = \text{Add}(\mathbf{P}_{t}, \mathbf{O}_{t+1})$
        %Observe and get vision features $\mathbf{V}_{t+1}$; 
    }
    \Return{$\mathbf{P}_{T}$;}
\end{algorithm}

% \begin{figure*}[t]
%   \centering
%   %\fbox{\rule{0pt}{2in} \rule{0.9\linewidth}{0pt}}
%     \includegraphics[width=0.95\linewidth]{IEEE-Transactions-LaTeX2e-templates-and-instructions-2/imgs/path.pdf}
%     \vspace{-2mm}
%    \caption{
%    % The paths of the same navigation task performed in an unseen environment are compared using ST-Booster and Bevbert.
%    Comparison of navigation paths for the same task performed in an unseen environment using ST-Booster and BEVBert~\cite{an2023bevbert}.
%    % In these two examples, we found that introducing ST-Booster can improve the agent's ability to perceive the environment, thereby avoiding passing by the key point without stopping and going in the opposite direction of the destination.
%    These examples demonstrate that incorporating ST-Booster enhances the agent's environmental perception, helping it avoid bypassing critical points without stopping and reducing instances of moving away from the destination.}
%    \vspace{-1mm}
%    \label{fig:navigation}
% \end{figure*}

% \vspace{-4mm}
% \begin{figure*}[t]
%   \centering
%   %\fbox{\rule{0pt}{2in} \rule{0.9\linewidth}{0pt}}
%    \includegraphics[width=0.98\linewidth]{IEEE-Transactions-LaTeX2e-templates-and-instructions-2/imgs/path_app_2.pdf}
%    \caption{Failed navigation examples with short and long instructions in unseen environments.}
%    \vspace{-4mm}
%    \label{fig:failed-path}
% \end{figure*}

\noindent1) Masked Language Modeling (MLM). 
This task is commonly used to improve the model's understanding of complex language structures~\cite{an2024etpnav}.
To be specific, the words of the input instructions are randomly masked with a probability of 15\%, prompting the model to predict the masked words based on contextual information. For the MLM task, the prediction loss for generating the masked words is defined by
\begin{linenomath*}
\begin{equation}
\mathcal{I}_{\textrm{MLM}}
= -\mathbb{E}_{(\mathbf{I}, \mathbf{\Omega}) \sim \mathcal{D}} \log 
\mathcal{P}_{\boldsymbol{\theta}}(\mathbf{I}_{\text{mask}} | \mathbf{I}_{\setminus \text{mask}}, \tilde{\mathbf{G}}^{\text{f}}_t, \tilde{\mathbf{M}}^{\text{f}}_t),
\end{equation}
\end{linenomath*}
where $\boldsymbol{\theta}$ denotes the policy parameters, $\mathbf{I}_{\text{mask}}$ and $\mathbf{I}_{\setminus \text{mask}}$ represent the masked and the unmasked portion of the instruction embeddings, respectively, and $\mathbf{\Omega}$ denotes the expert trajectory.

\noindent2) Hybrid Single Action Prediction (HSAP). 
To enhance the environmental perception and instruction comprehension by improving decision-making accuracy~\cite{an2023bevbert}, $\tilde{\mathbf{G}}^{\text{f}}_t$ and $\tilde{\mathbf{M}}^{\text{f}}_t$ are utilized to predict the long-term action $\mathbf{a}^{\text{G}}_t$ and short-term action $\mathbf{a}^{\text{M}}_t$, respectively. Then, we fuse them to get a robust decision. Specifically, if $\mathbf{a}^{\text{G}}_t$ is among the current candidate waypoint set $\mathcal{V}^\text{o}_{t, \text{cur}}$, we perform a weighted fusion of $\mathbf{a}^{\text{G}}_t$ and $\mathbf{a}^{\text{M}}_t$ with the weight $\gamma_t$; otherwise, we directly adopt $\mathbf{a}^{\text{G}}_t$, i.e.,
\begin{linenomath*}
\begin{equation}\label{eq:action_score}
\mathbf{a}_{t}^{\text{G}} = \textrm{FFN}(\tilde{\mathbf{G}}^{\text{f}}_t), \quad \mathbf{a}_{t}^{\text{M}} = \textrm{FFN}(\tilde{\mathbf{M}}^{\text{f}}_t),
%\vspace{-1mm}
\end{equation}
\end{linenomath*}

\begin{linenomath*}
\begin{equation}\label{eq:action_fuse}
\mathbf{a}_t =\left\{
\begin{aligned}
& \mathbf{\gamma}_t \mathbf{a}_{t}^{\text{G}} + (1-\mathbf{\gamma}_t) \mathbf{a}_{t}^{\text{M}},
~\textrm{if}~ \mathbf{a}_{t}^{\text{G}}\in \mathcal{V}^\text{o}_{t, \text{cur}}, \\
& \mathbf{a}_{t}^{\text{G}},~\text{otherwise}.
\end{aligned}
\right.
% \vspace{-1mm}
\end{equation}
\end{linenomath*}

After obtaining the fused actions, the HSAP loss at each single step is applied using the ground-truth target waypoints $\mathbf{a}^*_t$ from the expert trajectories as a reference, i.e.,
%\vspace{-3mm}
\begin{linenomath*}
\begin{equation}\label{eq:hsap}
%\vspace{-1mm}
\mathcal{L}_{\textrm{HSAP}} = -\mathbb{E}_{(\mathbf{L}, \mathbf{\Omega}, {\mathbf{a}_{t}^*})\sim \mathcal{D}}
\log \mathcal{P}_{\boldsymbol{\theta}}(\mathbf{a}_{t}^* | \mathbf{I}, \tilde{\mathbf{G}}^{\text{f}}_t, \tilde{\mathbf{M}}^{\text{f}}_t).
% \vspace{-1mm}
\end{equation}
\end{linenomath*}

\noindent3) Guided Attention Heatmap Prediction (GAHP).
This task is trained in a weakly supervised manner, with the Mean Squared Error (MSE) loss between the ground-truth GAH $\mathbf{H}_{t}^*$  and the predicted one $\mathbf{H}_{t}$, i.e.,

\begin{figure*}[!t]
  \centering
   \includegraphics[width=1\linewidth]{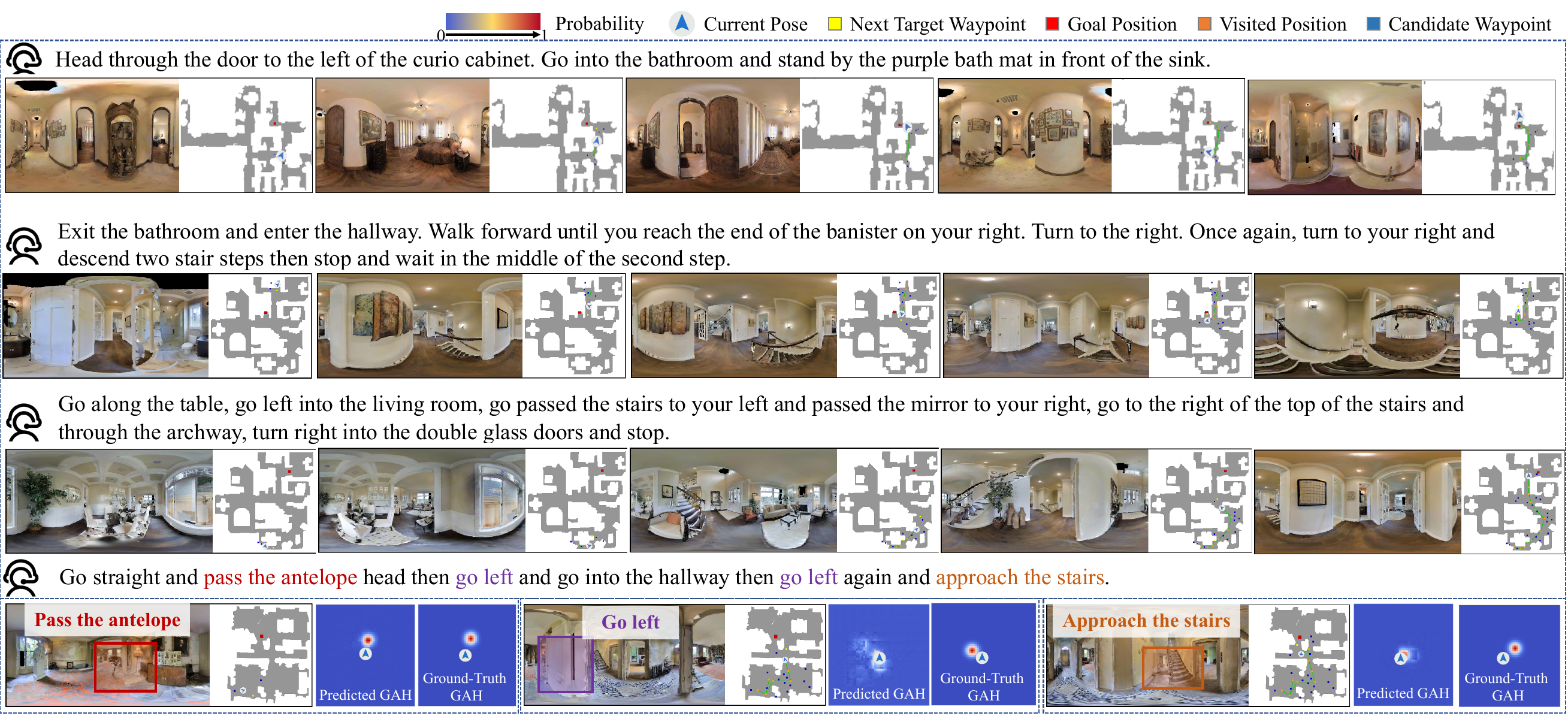}
   \caption{{Successful navigation examples under short and long instructions in unseen environments, along with visualizations of predicted and ground-truth GAHs along successful paths. The figure presents four representative cases. The top example illustrates navigation guided by a concise textual instruction (displayed above the navigation map), including the agent’s real-time panoramic observations and top-down views. The middle two examples depict navigation scenarios guided by longer instructions, while the bottom example shows the visualization of predicted and ground-truth GAHs along a successful trajectory.}\label{fig:navigation}}
   \vspace{-2mm}
\end{figure*}

\begin{linenomath*}
%\vspace{-3mm}
\begin{equation}\label{eq:gahp}
\mathcal{L}_{\textrm{GAHP}} = \mathbb{E}_{\mathbf{H}_{t}^*\sim \mathcal{D}}
(\textrm{MSE}(\mathbf{H}_{t}^*, \mathbf{H}_{t}) | \tilde{\mathbf{M}}^{\text{f}}_{t}).
%\vspace{-1mm}
\end{equation}
\end{linenomath*}

% {
% \begin{table}[t]
%   \centering
%   \caption{{Performance comparison on the val-unseen dataset when making decisions based on the predicted GAH maps}\label{tab:GAH-Performance}}
%   %
%   %\vspace{-3mm}
%   \resizebox{0.5\textwidth}{!}{
%      {\color{teal}
%     \begin{tabular}{l|ccccc}
%     \toprule
%     \multicolumn{1}{l|}{GAH Method}& TL    & NE ($\downarrow$)    & OSR ($\uparrow$)   & SR ($\uparrow$)    & SPL ($\uparrow$) \bigstrut\\
%     \midrule
%     Only-GAH & 7.15  & 9.14  & 3.92  & 2.88  & 1.89  \bigstrut[t]\\
%     WH+P-GAH & 20.58  & 8.51  & 27.35  & 19.74  & 10.12  \\
%     WH+F-GAH & 19.02  & 8.15  & 28.87  & 21.15  & 11.06  \bigstrut[b]\\
%     \bottomrule
%     \end{tabular}}%
%     }
%     %\vspace{-4mm}
  
% \end{table}%
% }

\noindent\textbf{Fine-Tuning.}
To prevent the model from overfitting to the offline expert data, we fine-tune the pre-trained model in the continuous simulation environments with the HSAP task. During this fine-tuning process, the topological and grid maps are updated in real-time based on the agent's observations. Additionally, we progressively transition from teacher-forcing to student-forcing during action reasoning phase~\cite{hong2022bridging} to balance learning stability and exploration. Specifically, in the teacher-forcing strategy, the agent navigates with ground-truth actions, enabling the model to quickly learn reasonable paths toward the target, while in the student-forcing strategy, the agent explores the unknown state-action space by navigating based on sampled results from the model's output 
This hybrid training approach effectively balances exploration and exploitation, allowing the model to learn expert behaviors effectively.

\section{Experiments}\label{experiment}
% \subsection{Experiment Settings}

This section presents the experimental setup, including the dataset, training details, and evaluation metrics, followed by comparisons between ST-Booster and SoTA baselines in VLN-CE. Extensive experiments further validate the contribution of each core module.

\subsection{Experimental Setup}
\label{sec:set}

\noindent\textbf{Datasets.}
% Discrete Vision-and-Language Navigation (VLN) tasks often utilize the Matterport3D (MP3D) environment~\cite{anderson2018vision}, comprising 90 scenes and 10,800 panoramic RGB-D images structured as navigation graphs.
The Matterport3D (MP3D) environment~\cite{anderson2018vision} with 90 scenes and 10,800 RGB-D images is commonly used for discrete VLN tasks. 
% To bridge the gap to a continuous setting, VLN-CE employs the Habitat simulator to reconstruct MP3D scenes in 3D, enabling seamless navigation in a continuous space. The VLN-CE dataset is derived by mapping the discrete Room-to-Room (R2R) dataset~\cite{anderson2018vision} into this continuous environment, and it is widely used for training and evaluating VLN-CE algorithms.
To enable continuous navigation, the VLN-CE benchmark reconstructs these scenes in 3D within the Habitat simulator~\cite{2019habitat}, adapting the Room-to-Room (R2R) dataset~\cite{anderson2018vision} for continuous environments.
 The dataset is split into train, val seen, val unseen, and test subsets. The val seen split involves new paths and instructions within familiar training environments, while val unseen and test contain paths in novel, unseen environments. 
 % To evaluate the proposed framework's effectiveness, models are typically tested on unseen environments to assess their generalization performance.
 To evaluate the proposed framework's generalization capabilities, models are typically assessed on unseen environments.

\begin{table*}[t]
  \centering
  % \caption{Performance comparison with state-of-the-art methods on the VLN-CE dataset.}
  \caption{Performance comparison of our proposed ST-Booster against state-of-the-art methods evaluated on the VLN-CE dataset. (P-GAH / FT-GAH: pretrained and fine-tuned GAH modules; EV+: evaluation using only GAH predictions; ST-FT / ST-EV: fine-tuning and evaluation stages of ST-Booster)}
  %\vspace{-2mm}
   \label{tab:performance-base}%
  \resizebox{1\textwidth}{!}{
    \begin{tabular}{l|ccccc|ccccc|ccccc}
    \toprule
    \multicolumn{1}{l|}{\multirow{2}[3]{*}{Method}} & \multicolumn{5}{c|}{Val Seen}         &       & \multicolumn{4}{c|}{Val Unseen} & \multicolumn{5}{c}{Test Unseen} \bigstrut\\
\cline{2-16}          & TL    & NE ($\downarrow$)    & OSR ($\uparrow$)   & SR ($\uparrow$)    & SPL ($\uparrow$)   & TL    & NE ($\downarrow$)    & OSR ($\uparrow$)   & SR ($\uparrow$)    & SPL ($\uparrow$)   & TL    & NE ($\downarrow$)    & OSR ($\uparrow$)   & SR ($\uparrow$)    & SPL ($\uparrow$) \bigstrut[t]\\ \midrule
    Seq2Seq~\cite{anderson2018vision} & 9.26 & 7.12  & 46    & 37    & 35    & 8.64      & 7.37  & 40    & 32    & 30    &  8.85     & 7.91  & 36    & 28    & 25 \\
    CM2~\cite{georgakis2022cross}   & 12.05  & 6.10  & 51    & 43    & 35    & 11.54  & 7.02  & 42    & 34    & 28    & 13.90  & 7.70  & 39    & 31    & 24 \\
    MGMap~\cite{chen2022weakly} & 10.12  & 5.65  & 52    & 47    & 43    & 10.00 & 6.28  & 48    & 39    & 34    & 12.30  & 7.11  & 45    & 35    & 28 \\
    CWP-RecBERT~\cite{hong2022bridging} & 12.50  & 5.02  & 59    & 50    & \underline{44}    & 12.23  & 5.74  & 53    & 44    & 39    & 13.31  & 5.89  & 51    & 42    & 36 \\
    Sim2Sim~\cite{krantz2022sim} & 11.18  & 4.67  & 61    & 52    & \underline{44}    & 10.69  & 6.07  & 52    & 43    & 36    & 11.43  & 6.17  & 52    & 44    & 37 \\
    ETPNav~\cite{an2024etpnav} & 11.78  & \underline{3.95}  & \underline{72}    & \underline{66}    & \textbf{59} & 11.99  & \textbf{4.71 } & 65    & 57    & 49    & 12.87  & 5.12  & 63    & 55    & 48 \\
    Ego2-Map~\cite{hong2023learning} & --     & --     & --     & --     & --     & --     & 4.94  & --     & 52    & 46    & 13.05  & 5.54  & 56    & 47    & 41 \\
    GridMM~\cite{wang2023gridmm} & 12.69  & 4.21  & 69    & 59    & 51    & 13.36  & 5.11  & 61    & 49    & 41    & 13.31  & 5.64  & 56    & 46    & 39 \\
    BEVBert~\cite{an2023bevbert} & --     & --     & --     & --     & --     & --     & 4.57  & \underline{67}    & \underline{59}    & \underline{50} & --     & \textbf{4.70 } & \textbf{67} & \textbf{59} & \textbf{50} \\
    Scale-VLN~\cite{wang2023scale} & --     & --     & --     & --     & --     & --     & 4.80  & --     & 55    & \textbf{51}    & --     & 5.11  & --     & 55    & \textbf{50} \bigstrut[b]\\
    \hline
    \rowcolor{rowcolor}
    ST-Booster (\textbf{Ours}) & 13.29  & \textbf{3.71 } & \textbf{77} & \textbf{69} & \textbf{59} & 13.14  & \underline{4.77}  & \textbf{68} & \textbf{61} & \underline{50} & 13.42  & \underline{4.83}  & \underline{66}    & \textbf{59} & \textbf{50} \bigstrut\\
    \bottomrule
    
    \end{tabular}}%
 
\end{table*}%

\begin{table}[ht]
  \centering
  \caption{Ablation study on the MGAF and VGWG modules Evaluated on Varying Map Experts}
  \label{tab:alation1}%
  \resizebox{0.47\textwidth}{!}{
    \begin{tabular}{l|cc|ccccc}
    \toprule
    \multicolumn{1}{l|}{Expert} & MGAF & VGWG   & TL    & NE ($\downarrow$)    & OSR ($\uparrow$)   & SR ($\uparrow$)    & SPL ($\uparrow$) \bigstrut\\
    \midrule
    \multirow{4}[2]{*}{Topological Map} & \ding{55} & \ding{55} & 13.87  & 5.24  & 61.6  & 53.6  & 43.1  \bigstrut[t]\\
          & \ding{55} & \ding{51} & 13.15  & 5.38  & 60.3  & 54.1  & 43.8  \\
          & \ding{51} & \ding{55} & 13.46  & \underline{5.12}  & \underline{64.4}  & \underline{55.5}  & \underline{45.2}  \\
          & \ding{51} & \ding{51} & 13.15  & \textbf{5.02}  & \textbf{64.8}  & \textbf{56.9}  & \textbf{46.6}  \bigstrut[b]\\
    \hline
    \multirow{4}[2]{*}{Grid Map} & \ding{55} & \ding{55} & 9.01  & \textbf{5.65}  & 46.3  & 38.4  & 33.3  \bigstrut[t]\\
          & \ding{55} & \ding{51} & 8.64  & \underline{5.76}  & 44.8  & 38.6  & 33.6  \\
          & \ding{51} & \ding{55} & 10.53  & 5.77  & \underline{51.4}  & \underline{41.8}  & \underline{35.0}  \\
          & \ding{51} & \ding{51} & 10.50  & 5.70  & \textbf{52.1}  & \textbf{43.0}  & \textbf{36.2}  \bigstrut[b]\\
    \hline
    \multirow{4}[2]{*}{Hybrid Map} & \ding{55} & \ding{55} & --  &   \underline{4.57} & 67.0  & 59.0  & 50.0  \bigstrut[t]\\
          & \ding{55} & \ding{51} & 13.05  & \textbf{4.52}  & \underline{67.2}  & 59.7  & \underline{49.7}  \\
          & \ding{51} & \ding{55} & 13.22  & 4.87  & 66.7  & \underline{59.8}  & 49.3  \\
          & \ding{51} & \ding{51} & 13.14   & 4.77  & \textbf{67.6} & \textbf{61.0} & \textbf{50.2} \bigstrut[b]\\
    \bottomrule
    \end{tabular}}
    \vspace{-3mm}
\end{table}%

\noindent\textbf{Implementation Details.}
% We leverage ViT-B/16-CLIP~\cite{radford2021learning} to extract visual features, where the feature dimension is 768.
% For grid-based representations, the extracted features are downsampled to a scale of $14\times 14$. The corresponding grid map covers an area of $11\text{m}\times 11\text{m}$, with each grid cell representing a 1$\text{m}^2$.
% During pre-training, we initialize our model with pre-trained LXMERT~\cite{tan2019lxmert} and conduct 200,000 iterations of offline training using a batch size of 64. The model that achieves the highest performance on pre-training metrics is then subjected to fine-tuning through an additional 20,000 iterations of online training, with a reduced batch size of 16. Ultimately, the fine-tuned model is evaluated in previously unseen environments, with the highest-performing variant selected for further analysis.
The visual feature extraction of ST-Booster is based on ViT-B/16-CLIP~\cite{radford2021learning}, which generates representations with a dimension of~768.
For grid-based spatial representations, the extracted features are downsampled to match the encoder patch size $P\times P=$14$\times$14. When the spatial dimensions are set to $U=V=11$ with a cell resolution of 1\,m, the grid map covers an area of $11\,\text{m}\times11\,\text{m}$, where each cell corresponds to 1\,$\text{m}^2$. With an upsampling factor of $m=n=5$, the GAH map expands to $55\,\text{m}\times 55\,\text{m}$, and each cell has a side length of 0.2\,m. 
During pretraining, our model is initialized with pre-trained LXMERT weights~\cite{tan2019lxmert} and undergoes 200,000 iterations of offline training using a batch size of~64. The model with the highest pretraining performance is subsequently fine-tuned through an additional 20,000 iterations of online training with a batch size of 16. Finally, the fine-tuned model is evaluated on unseen environments, and the highest-performing variant is selected for further analysis.

\begin{figure*}[ht]
  \centering
   \includegraphics[width=1\linewidth]{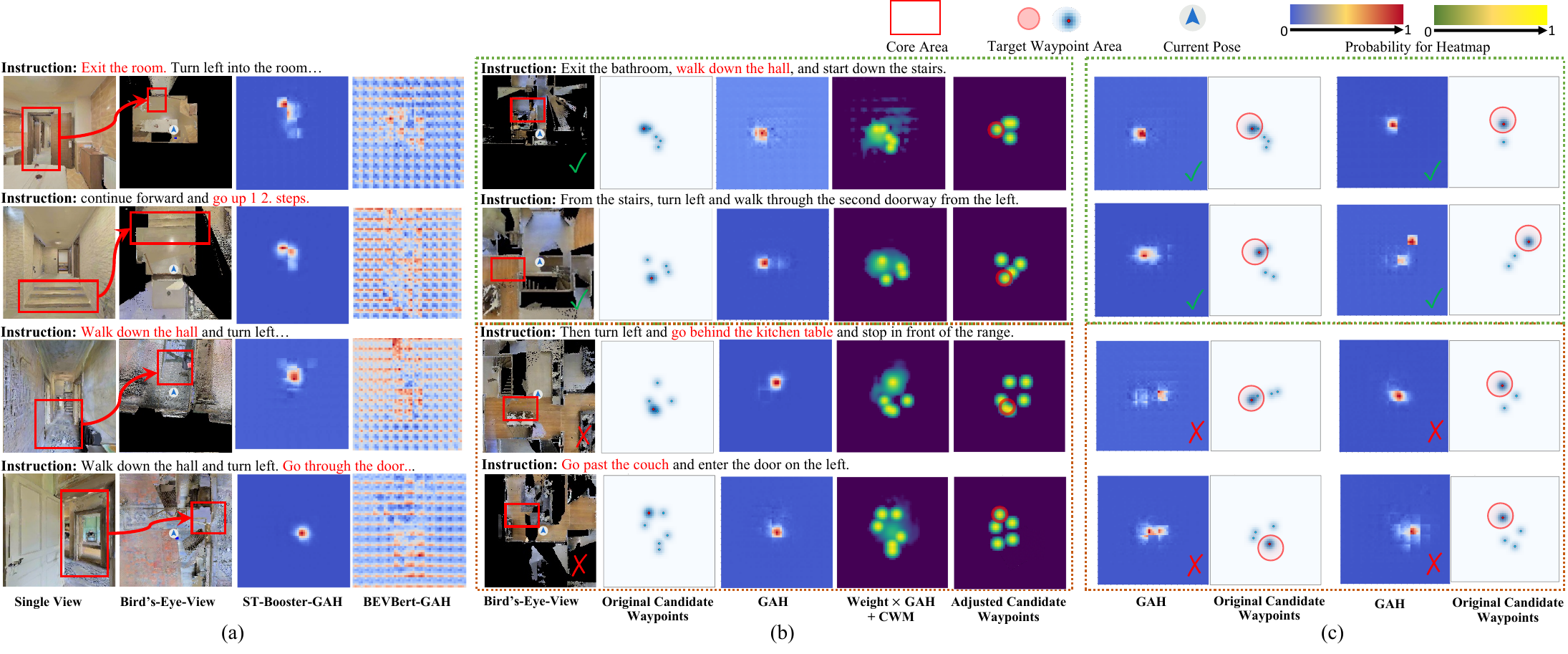}
   \caption{{Visualization of GAH generator predictions. (a) Generated GAHs of BEVBert and ST-Booster on val unseen split of VLN-CE. 
(b) The influence of GAH on candidate waypoint distributions under correct and erroneous predictions.
(c) Successful and failed examples generated by GAH.}\label{fig:GAH}}
   
\end{figure*}

\noindent\textbf{Evaluation Metrics.}
To evaluate the performance of the proposed model, we utilized several standard metrics, including (i) Trajectory Length (TL), measuring the average path length; (ii) Navigation Error (NE), representing the average distance between the agent's final position and the target; (iii) Oracle Success Rate (OSR), the rate at which the agent reaches a position within $3\,\text{m}$ of the target; (iv) Success Rate (SR), the rate at which the agent stops within $3\,\text{m}$ of the target; and (v) Success rate weighted by Path Length (SPL), which balances success rate with path efficiency. These metrics provide a comprehensive assessment of the model's navigation performance on both seen and unseen splits.

\subsection{Comparison With the State-of-the-Art Methods}
\label{sec:sota}

We compare ST-Booster with several state-of-the-art methods listed in \Cref{tab:performance-base}. Sim2Sim~\cite{krantz2022sim} and CWP-RecBERT~\cite{hong2022bridging} adopt end-to-end frameworks that jointly model visual features and historical states, while ETPNav~\cite{an2024etpnav} encodes topological maps to capture navigation patterns, and GridMM~\cite{wang2023gridmm} applies grid-based representations for fine-grained spatial reasoning. BEVBert~\cite{an2023bevbert}, the most comparable hybrid-map approach, trains global and local maps separately and fuses them at the decision stage. In contrast, ST-Booster performs feature-level hybrid map fusion, jointly optimizing waypoint generation and navigation reasoning through enhanced spatiotemporal perception. As illustrated in \Cref{tab:performance-base}, ST-Booster surpasses BEVBert by 2\% in SR on the val-unseen split and achieves the best SR (59\%) and SPL (50\%) on the test-unseen split. Visualization results in \Cref{fig:navigation} further demonstrate that by improving multi-modal hybrid map perception, ST-Booster more effectively captures spatiotemporal relationships and achieves higher navigation success in unseen environments.

\begin{table}[t]
  \centering
  \caption{{Performance Comparison of GAH Models at Different Training Stages on the Val-Unseen Split.
(P-GAH and FT-GAH denote the GAH prediction modules obtained after pretraining and fine-tuning, respectively; EV+ indicates decision-making using only GAH predictions during evaluation; ST-FT and ST-EV represent the fine-tuning and evaluation stages of ST-Booster, respectively)}\label{tab:GAH-Use}}
  %\vspace{-3mm}
  
  %\resizebox{0.5\textwidth}{!}{
   %{\color{teal}
    \begin{tabular}{l|ccccc}
    \toprule
    \multicolumn{1}{l|}{Method with GAH}& TL    & NE ($\downarrow$)    & OSR ($\uparrow$)   & SR ($\uparrow$) & SPL ($\uparrow$) \bigstrut\\
    \midrule
    EV+P-GAH & 20.58 & 8.51 & 27.35 & 19.74 &10.12 \bigstrut[t] \\
    EV+FT-GAH & 19.02 & 8.15 & 28.87 & 21.15 & 11.06 \\
    ST-FT+P-GAH & 16.51  & 4.79  & 68.03  & 58.08 & 45.03  \\
    ST-FT+FT-GAH & 15.18 & 4.74  & 65.85  & 58.78  & 47.67  \\
    ST-EV+P-GAH &  13.14 &  4.77 &  67.50 & 60.95 & 50.16  \\
    ST-EV+FT-GAH & 13.02  & 4.72  & 67.54  & 61.12 & 50.55 
     
    \bigstrut[b]\\
    \bottomrule
    \end{tabular}%
    %}}
    \vspace{-3mm}
  
\end{table}% 

\begin{figure}[!t]
  \centering
   \includegraphics[width=1\linewidth]{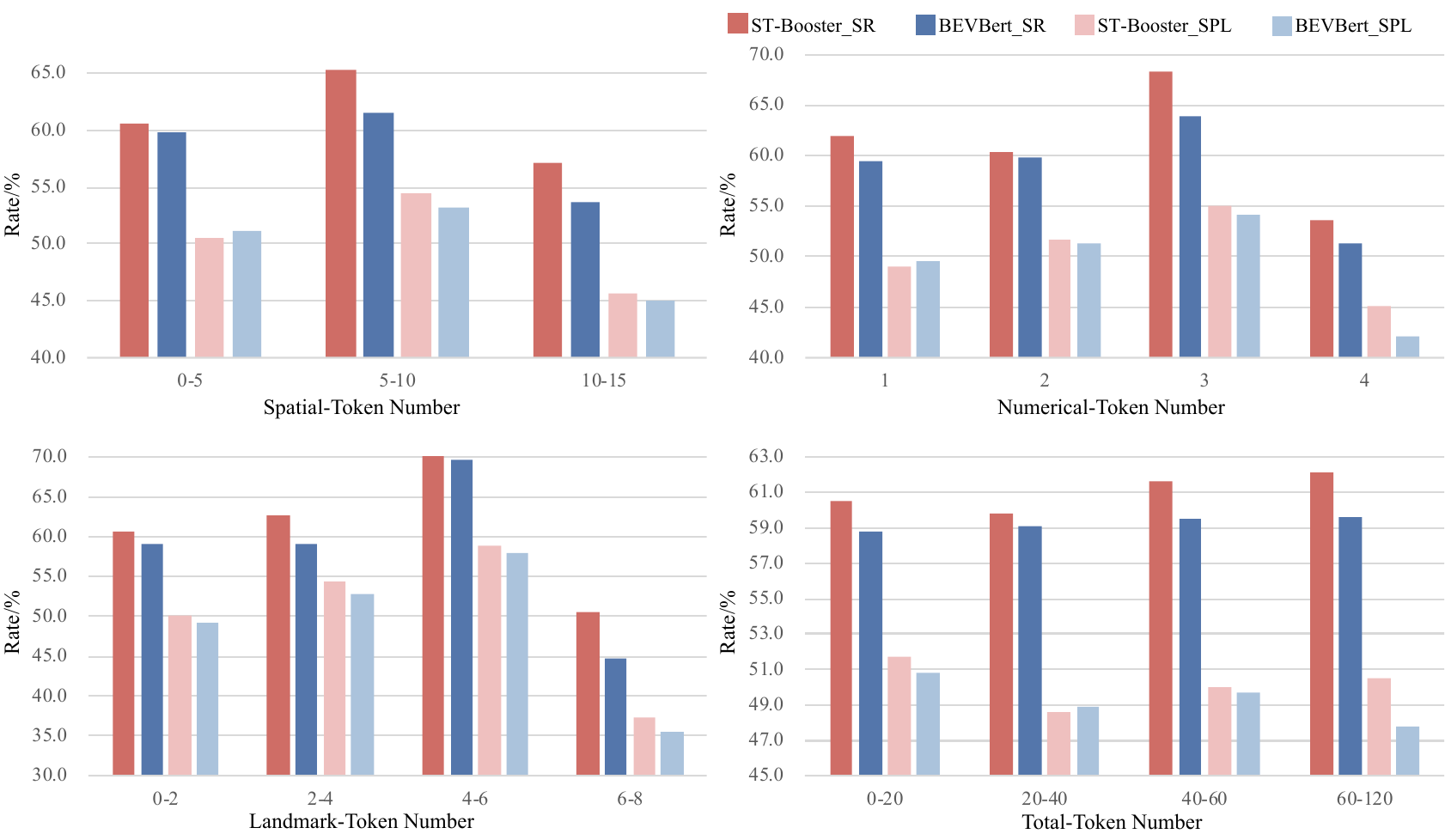}
   %\vspace{-6mm}
   \caption{Comparison of navigation performance with respect to task complexity in VLN-CE unseen split. Experimental results demonstrate that the performance of ST-Booster is consistently superior to the selected SOTA methods across all complexity dimensions.}
   \vspace{-4mm}
   \label{fig:ins}
\end{figure}

\begin{figure*}[ht]
  \centering
   \includegraphics[width=1.\linewidth]{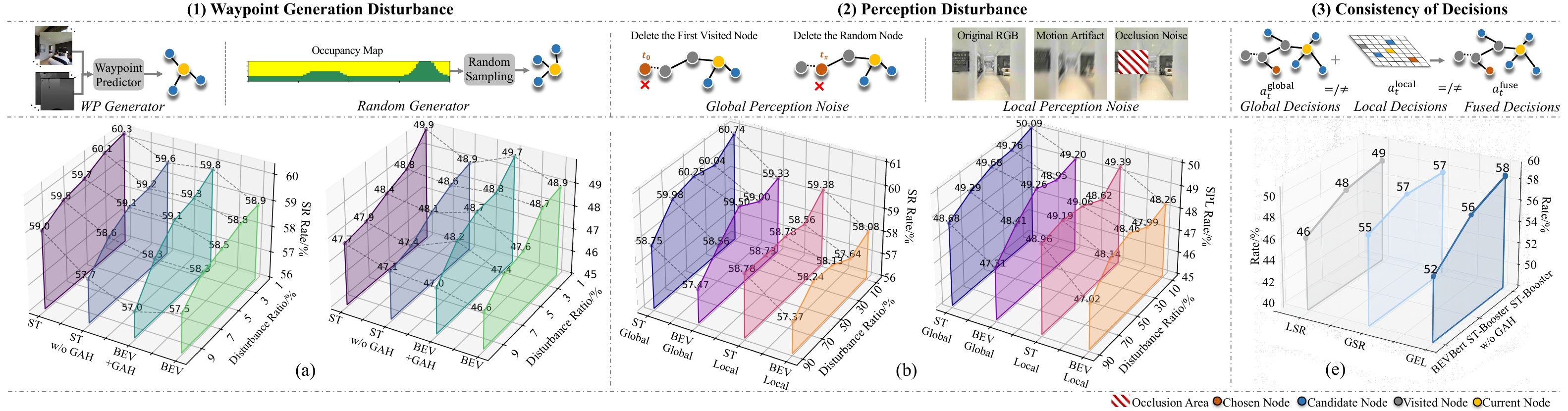}
   \caption{{Robustness analysis of different VLN-CE agents, ST-Booster (ST) and BEVBert (BEV), under various disturbances. (a) SR and path efficiency SPL under different waypoint disturbance ratios. `BEV+GAH' denotes BEVBert with the VGWG module generating candidate waypoints from GAHs. (b) SR and SPL under different perception disturbance levels, where `VLN-CE Global' and `VLN-CE Local' indicate agents affected by global and local perception disturbances, respectively. (c) Decision consistency results, where LSR and GSR denote the SR of local and global experts, and GEL indicates the agreement probability between their decisions.}\label{fig:disturb}}
   \vspace{-2mm}
\end{figure*}

\subsection{Ablation Study}
\label{sec:ablation}
%This ablation study aims to assess the effectiveness of ST-Booster with or without GAH.
%As shown in \Cref{tab:alation1}, among various map expert strategies, the hybrid map expert strategy yields superior navigation performance. In addition, the combination of ST-Booster and GAH consistently improves performance across all map expert strategies. When using topology and grid maps for decision-making individually, the proposed method results in a 3.3\% and 4.6\% increase in SR, respectively, while for the hybrid map expert, the success rate improves by 2\%.
%These findings demonstrate that ST-Booster effectively enhances the agent's environmental understanding, while GAH provides valuable prior knowledge for action reasoning. The proposed method plays a critical role in improving the navigation performance across different map expert strategies.
This ablation study aims to validate the synergistic optimization effects of the feature fusion capability of the MGAF module and the instruction-guided waypoint prediction mechanism of the VGWG module. As illustrated in \Cref{tab:alation1}, we consider three different map experts to predict actions with or without the key modules proposed in ST-Booster, including topological, grid, and hybrid map experts. Specifically, when activating MGAF without VGWG, it improves the success rate (SR) of the topological map expert by 1.9\% and the grid map expert by 3.4\%. With VGWG enabled, MGAF further boosts these improvements to 2.8\% and 4.6\% respectively, while achieving a 2.0\% SR enhancement for the hybrid map expert. These improvements confirm MGAF's effectiveness in facilitating complementary fusion of dual-map perceptions and strengthening instruction-feature alignment, thereby benefiting both individual and hybrid experts. Furthermore, the VGWG module demonstrates its distinct value through instruction-aware waypoint distribution prediction, which focuses attention on navigation-critical regions, contributing an additional 1.2\% SR gain to the hybrid expert. The combined application of both modules enables the hybrid expert to achieve optimal performance with 61.0\% SR, illustrating the effectiveness of the modules within ST-Booster.

\subsection{Performance Analysis}
\label{sec:performance}

\noindent\textbf{GAH Performance.}
{To comprehensively assess the role of the GAH, we analyze GAH via visualization, waypoint influence, decision ability, and train–test usage.}

\noindent\textit{GAH Visualization Analysis.}
{Compared with GAHs from baseline image features, BEVBert lacks global cues and produces checkerboard heatmaps (\Cref{fig:GAH}(a)). With enhanced representation, ST-Booster yields smoother GAHs that highlight instruction-relevant landmarks. GAH usually steers candidates toward goal regions, making the fused distribution closer to the ground truth (illustrated in the top two rows of \Cref{fig:GAH}(b) and \Cref{fig:GAH}(c)), but its coarse cues add noise in some cases (illustrated in the bottom two rows of \Cref{fig:GAH}(b) and \Cref{fig:GAH}(c)).}

\noindent\textit{GAH Decision Capability Analysis.}
{We quantitatively evaluate GAH’s decision capability. As illustrated in \Cref{tab:GAH-Use}, weighting the pretrained GAH with the candidate waypoint distribution and selecting the highest-value waypoint yields a SR of 19.74\%. After fine-tuning VGWG with the mean squared error between the GAH and the ground-truth heatmap, SR rises to 21.15\%. Although GAH effectively highlights instruction-relevant regions, its SR remains lower than when combined with the dual-map expert. The gap mainly stems from prediction errors and the lack of backtracking. Therefore, GAH used as an auxiliary prior to complement spatial information improves the robustness of the dual-map experts.}

\noindent\textit{GAH Performance Improvement Analysis.}
{We investigate when to apply GAH. 
As illustrated in \Cref{tab:GAH-Use}, using GAH during fine-tuning reduces SR to 58.78\% due to over-concentrated candidates and reduced exploration. In contrast, applying GAH only during evaluation improves SR to 61.12\%, as the fine-tuned dual-map experts tolerate imperfect GAHs and focus on critical regions using prior knowledge. Overall, GAH may constrain training exploration but improves inference, so using it as an auxiliary module at evaluation works best.
}

\noindent\textbf{Consistency Against Task Complexity.}
We evaluate the generalization performance of ST-Booster in complex tasks within unseen environments, conducting a statistical analysis of the proposed framework’s performance. Task complexity is quantified based on the instruction’s complexity across four dimensions. For instance, we count the number of terms representing spatial elements and landmarks in each instruction, where a higher count indicates increased spatial complexity of the navigation task. Similarly, we analyze the frequency of numeric terms and the total vocabulary count in each instruction, reflecting the task's cognitive difficulty in handling long-term dependencies. Experimental results in \Cref{fig:ins} demonstrate that the performance of ST-Booster is consistently superior to the selected SOTA methods across all complexity dimensions. This confirms that ST-Booster effectively enhances complex spatiotemporal perception and improves navigation performance.

\noindent\textbf{Robustness from Spatiotemporal Complementarity.}
{We evaluate the model’s robustness by testing the complementary effects of spatiotemporal perception under three controlled disturbances: (1) field-of-view loss, simulated by randomly selecting candidate waypoints to disrupt the uniformity of the original waypoint distribution; (2) local noise, introduced by applying motion blur and partial masking to local observations to mimic real camera disturbances; and (3) long-term memory decay, simulated by randomly removing previously visited waypoints to emulate memory limitations.
As illustrated in \Cref{fig:disturb}, our model consistently outperforms BEVBert in SR and SPL across all conditions and shows smaller performance drops under unreliable perception. This robustness arises from the spatiotemporal complementarity between the dual-map experts. When short-term local observations become noisy, the topological expert offers long-term semantic stability. Conversely, when global topology weakens, the grid expert preserves spatial accuracy through real-time sensing. Their interaction at the feature level allows information to flow and adjust across time and space, maintaining stable navigation under various disturbances.}

\noindent\textbf{Consistency from Spatiotemporal Alignment.}
{We evaluate decision consistency achieved through spatiotemporal alignment between the two experts. In BEVBert, the grid and topological experts act independently and are fused only at the decision stage, leading to temporal and semantic misalignment. As illustrated in \Cref{fig:disturb}(c), our model enables bidirectional spatiotemporal feature fusion, aligning local observations with global semantics across time and space. This alignment improves information integration, combining short-term perception with long-term memory for more coherent decisions. In unseen environments, it increases SR of the grid and topological experts to 49\% and 57\%, respectively, and raises decision agreement to 58\%, a 6\% gain over BEVBert, confirming more consistent and reliable navigation.

}

\begin{table}[t]
  \centering
  \caption{The effect of grid map size on navigation performance}
  \label{tab:effect-map}%
  \resizebox{0.47\textwidth}{!}{
    \begin{tabular}{l|c|ccccc}
    \toprule
    \multicolumn{1}{l|}{Scale Size} & Cell Resolution  & TL    & NE ($\downarrow$)    & OSR ($\uparrow$)   & SR ($\uparrow$)    & SPL ($\uparrow$) \bigstrut\\
    \midrule
    \multirow{2}[2]{*}{7$\times$ 7} & 0.5\,m  & 14.02  & 5.10  & 63.9  & 55.6  & 45.6  \bigstrut[t]\\
          & 1\,m     & 12.73  & 4.80  & 63.9  & 57.2  & 47.9  \bigstrut[b]\\
    \hline
    \multirow{2}[2]{*}{11$\times$11} & 0.5\,m  & 13.05  & 4.99  & 65.7  & 57.9  & 49.2  \bigstrut[t]\\
          & 1\,m    & 13.14  & 4.77  & 67.6  & 61.0  & 50.2  \bigstrut[b]\\
    \hline
    \multirow{2}[2]{*}{15$\times$15} & 0.5\,m    & 12.70  & 4.81  & 66.2  & 58.7  & 49.9 \bigstrut[t]\\
          & 1\,m    & 13.2  & 4.78  & 66.6  & 59.8  & 48.8  \bigstrut[b]\\

    \bottomrule
    \end{tabular}}%
    \vspace{-4mm}
  
\end{table}%

\vspace{-4mm}
\subsection{Influence of Design Parameters}
\label{sec:para}

\noindent\textbf{Grid Map Size.}
We analyze the effect of grid map size on navigation performance in the val unseen split of VLN-CE. We evaluate grid maps with sizes \{7$\times$7, 11$\times$11, 15$\times$15\} and cell resolutions of \{$0.5\,\text{m}$, $1\,\text{m}$\} for grid maps. As illustrated in \Cref{tab:effect-map}, increasing the cell size generally improves navigation performance. This improvement is caused by the fact that smaller cells capture more noise and redundant details, interfering with the agent's extraction of key environmental cues. Furthermore, when the cell resolution is $0.5\,\text{m}$, increasing the map size boosts success rates by incorporating more environmental information. However, with a $1\,\text{m}$ cell resolution, the performance of the 15$\times$15 map declines.
This is due to the excessive map size, which results in an increased region overlap between adjacent time steps. This causes redundant computations and data processing complexity. The results indicate that balancing scale size and cell resolution is crucial for optimizing navigation. 
\begin{table}[t]
  \centering
  \caption{{Performance comparison on the val-unseen dataset between random noised maps (RNMs) and GAHs under different map weight settings}\label{tab:effect-weight}}
  
  %\resizebox{0.5\textwidth}{!}{
    %{\color{teal} % ← 整个表格内容变色
    \begin{tabular}{c|c|ccccc}
      \toprule
      Map & GAH Weight & TL & NE ($\downarrow$) & OSR ($\uparrow$) & SR ($\uparrow$) & SPL ($\uparrow$) \\
      \midrule
      \multirow{3}[2]{*}{RNM}
        & $1.0\times10^{-4}$ & 13.12 & 5.0 & 65.3 & 56.9 & 47.2 \bigstrut[t] \\
        & $1.0\times10^{-5}$ & 13.13 & 4.8 & 66.7 & 58.4 & 48.6 \\
        & $1.0\times10^{-6}$ & 13.18 & 4.9 & 66.9 & 59.6 & 49.6 \bigstrut[b] \\
      \hline
      \multirow{3}[2]{*}{GAH}
        & $1.0\times10^{-4}$ & 13.48 & 5.0 & 65.2 & 57.7 & 47.2 \bigstrut[t] \\
        & $1.0\times10^{-5}$ & 13.14 & 4.8 & 67.6 & 61.0 & 50.2 \\
        & $1.0\times10^{-6}$ & 13.13 & 4.8 & 67.1 & 59.8 & 49.6 \bigstrut[b] \\
      \bottomrule
    \end{tabular}
    %}} % ← 颜色作用域结束
  
\vspace{-5mm} 
\end{table}

\noindent\textbf{GAH Weights.}
% As a coarse-grained expert prior, GAH is applied by multiplying it with a weight and adding it to the probability heatmap predicted by the original waypoint predictor. This experiment analyzes the sensitivity of performance on val unseen split dataset to different GAH weights. We tested three weight values, as shown in \Cref{tab:effect-weight}. Reducing the GAH weight from \(10^{-4}\) to \(10^{-5}\) improves most metrics. However, further reduction to \(10^{-6}\) slightly degrades performance. This indicates that while GAH provides useful prior knowledge, its coarse-grained nature can introduce bias limit exploration, and reduce performance. In this paper, we use a GAH weight of \(10^{-5}\) with the best performance for the remaining experiments.
{We evaluate the effect of different GAH fusion weights to assess the contribution of GAH to navigation reasoning. As a coarse-grained expert prior, the GAH is fused with the predicted waypoint heatmap $\mathbf{P}_t$ via weighted addition $(\delta)$. The weight selection is first driven by the magnitude gap between $\mathbf{H}_t$ and $\mathbf{P}_t$. The GAH follows the scale of the ground-truth map $\mathbf{H}^{*}_t$, obtained by converting the next waypoint into a Gaussian-blurred heatmap in $[0,1]$ and applying a scaling factor $\rho = 10$:
$\mathbf{H}^*_t = \rho \cdot \frac{g_i}{g^{\max}}$.
As a result, $\max{\mathbf{H}_t} = 10$. In contrast, $\mathbf{P}_t$ is a softmax-normalized distribution with a total sum of $1$ and an average of approximately $7 \times 10^{-4}$. This substantial scale mismatch requires using a small fusion weight for $\mathbf{H}_t$. Moreover, large weights make the fused waypoint distribution overly concentrated, amplifying prediction errors from $\mathbf{H}_t$ and degrading performance. Consistent with this analysis, experiments show that reducing the GAH weight from $10^{-4}$ to $10^{-5}$ yields performance improvements across most metrics, whereas further decreasing it to $10^{-6}$ results in a slight drop.}

{
We compare GAH $\mathbf{H}_t$ with a Random Noise Map (RNM) having the same order of magnitude. At both $10^{-4}$ and $10^{-5}$, RNM performs worse than $\mathbf{H}_t$, with a 2.6\% lower success rate at $10^{-5}$. This indicates that RNM changes waypoint distributions but fails to guide sampling towards task-relevant regions, while $\mathbf{H}_t$ effectively highlights relevant areas, improving waypoint inference. When the weight is reduced to $10^{-6}$, the performance gap narrows, showing that such a small weight has minimal impact on waypoint distribution. Overall, an appropriate weight provides balanced prior knowledge, while an excessively large weight may introduce bias and limit exploration. Based on these results, we adopt a GAH weight of $10^{-5}$ for optimal performance in subsequent experiments.}
% \subsubsection{Parameter Analysis}

% \begin{table}[t]
%   \centering
%   \caption{The effect of GAH weight on navigation performance}
%   %\vspace{-3mm}
%   \resizebox{0.42\textwidth}{!}{
%     \begin{tabular}{l|ccccc}
%     \toprule
%     \multicolumn{1}{l|}{GAH Weight}& TL    & NE ($\downarrow$)    & OSR ($\uparrow$)   & SR ($\uparrow$)    & SPL ($\uparrow$) \bigstrut\\
%     \midrule
%     $1.0\times10^{-4}$ & 13.62  & 4.95  & 65.2  & 57.7  & 47.2  \bigstrut[t]\\
%     $1.0\times10^{-5}$ & 13.14  & 4.77  & 67.6  & 61.0  & 50.2  \\
%     $1.0\times10^{-6}$ & 13.17  & 4.83  & 67.1  & 59.8  & 49.6  \bigstrut[b]\\
%     \bottomrule
%     \end{tabular}}%
%     %\vspace{-4mm}
%   \label{tab:effect-weight}%
% \end{table}%

\section{Conclusion}\label{conclusion}
To tackle the perception challenges in VLN-CE, this paper proposes an iterative spatiotemporal enhancement method named ST-Booster, which encompasses three interconnected modules.
The hierarchical spatiotemporal encoding (HSTE) module decouples environmental understanding into long-term navigation memory encoded in topological graphs and short-term geometric details captured by grid maps. The multi-granularity alignment fusion (MGAF) module introduces the bidirectional geometric-aware transformation to enable semantic-aware knowledge exchange across spatiotemporal scales. The value-guided waypoint generation (VGWG) module bridges perceptual features with actionable navigation priors. Extensive comparative experiments and comprehensive performance analysis are conducted, demonstrating that our proposed method outperforms the state-of-the-art approaches, particularly in complex tasks with environmental disturbances. Future work will extend this paradigm to tackle open-world perception challenges, including partial observability and lifelong map maintenance, further bridging the gap between virtual benchmarks and real-world deployment.


\begin{nolinenumbers}
\bibliographystyle{IEEEtran} % use IEEEtran.bst style
\bibliography{reference.bib}
\end{nolinenumbers}

\vfill

\end{document}